\documentclass[runningheads]{llncs}
\usepackage{graphicx}
\usepackage{tikz}
\usepackage{comment}
\usepackage{amsmath,amssymb} 
\usepackage{color}

\usepackage{bbm}
\usepackage{booktabs}
\usepackage{makecell}
\usepackage{algorithm}
\usepackage{algorithmic}
\usepackage[accsupp]{axessibility}  
\usepackage{tabu}
\usepackage{booktabs}
\usepackage{multirow}
\usepackage{subfigure}
\usepackage{blindtext}
\usepackage{multicol}
\usepackage{color}

\begin{document}
\pagestyle{headings}
\mainmatter
\def\ECCVSubNumber{6739}  

\title{
UIA-ViT: Unsupervised Inconsistency-Aware Method based on Vision Transformer \\for Face Forgery Detection 
} 

\titlerunning{Unsupervised Inconsistency-Aware Method for Face Forgery Detection }
%
\author{Wanyi Zhuang\and Qi Chu\inst{\ast}\and Zhentao Tan\and Qiankun Liu\and Haojie Yuan\and \\Changtao Miao\and Zixiang Luo \and Nenghai Yu}
\authorrunning{Wanyi Zhuang et al.}
%
\institute{CAS Key Laboratory of Electromagnetic Space Information, \\University of Science and Technology of China \\
\email{wy970824@mail.ustc.edu.cn, qchu@ustc.edu.cn, \{tzt,liuqk3,doubihj, miaoct,zxluo\}@mail.ustc.edu.cn, ynh@ustc.edu.cn.}
}
\maketitle

\renewcommand{\thefootnote}{\fnsymbol{footnote}}
\footnotetext[1]{Corresponding authors.} 

\begin{abstract}
Intra-frame inconsistency has been proved to be effective for the generalization of face forgery detection. However, learning to focus on these inconsistency requires extra pixel-level forged location annotations. Acquiring such annotations is non-trivial. Some existing methods generate large-scale synthesized data with location annotations, 
which is only composed of real images and cannot capture the properties of forgery regions.
Others generate forgery location labels by subtracting paired real and fake images, yet such paired data is difficult to collected and the generated label is usually discontinuous. 
To overcome these limitations, we propose a novel Unsupervised Inconsistency-Aware method based on Vision Transformer, called UIA-ViT, which only makes use of video-level labels and can learn inconsistency-aware feature without pixel-level annotations.
Due to the self-attention mechanism, the attention map among patch embeddings naturally represents the consistency relation, making the vision Transformer suitable for the consistency representation learning.
Based on vision Transformer, we propose two key components: Unsupervised Patch Consistency Learning (UPCL) and Progressive Consistency Weighted Assemble (PCWA). 
UPCL is designed for learning the consistency-related representation with progressive optimized pseudo annotations.
PCWA enhances the final classification embedding with previous patch embeddings optimized by UPCL to further improve the detection performance. Extensive experiments demonstrate the effectiveness of the proposed method.

\end{abstract}

\section{Introduction}
Face forgery technologies\cite{FaceSwap2019,DeepFakes2019,thies2016face2face} have been greatly promoted with the development of image generation and manipulation. The forged facial images can even deceive human beings, which may be abused for malicious purposes, leading to serious security and privacy concerns, e.g. fake news and evidence. Thus, it's of great significance to develop powerful techniques to detect fake faces.

Early face forgery detection methods\cite{chollet2017xception,nguyen2019capsule,wang2020cnn} regard this task as a binary classification problem and achieve admirable performance in the intra-dataset detection with the help of deep neural networks. However, they fail easily when generalizing to other unseen forgery datasets where the identities, manipulation types, compression rate, \textit{etc.} are quite different. To improve the generalization of detection, common forged artifacts or inconsistency produced by face manipulation techniques are explored by recent methods, such as eye blinking frequency \cite{li2018ictu}, affine warping \cite{li2018exposing}, image blending \cite{li2020face}, temporal inconsistency \cite{zheng2021exploring,sun2021dual}, intra-frame inconsistency \cite{PCL,chen2021local} and so on. Among them, intra-frame inconsistency has been proved to be able to effectively improve the generalization performance of the detection, since the common face forgery strategy (manipulation and blending) causes the inconsistency between the forged region and the original background. However, learning to focus on these inconsistency requires extra pixel-level forged location annotations. Acquiring such annotations is non-trivial.  Generating the large-scale synthesized data (e.g. simulated stitched images\cite{PCL}) with pixel-level forged location annotations seems to be an intuitive solution. 
Although it can produce accurate pixel-level location annotations, models can not capture the properties of forgery regions, since the generated data is only composed of real images.
Other works \cite{chen2021local,sun2021dual} attempt to generate annotated forged location labels by subtracting forgery image with its corresponding real image. However, these paired images are usually unavailable, especially in the real-world scenes. Even though such paired data can be collected, the forgery region annotations tend to be discontinuous and inaccurate, which are sub-optimal for intra-frame consistency supervision. Therefore, we propose an unsupervised inconsistency-aware method that extracts intra-frame inconsistency cues without pixel-level forged location annotations.

The key of unsupervised inconsistency-aware learning is how to realize forgery location estimation. In this paper, we apply the widely used multivariate Gaussian estimation (MVG)\cite{rippel2021mvg,li2021cutpaste} to represent the real/fake features and generate pseudo annotations through it. Based on this idea, we can force the model to focus on intra-inconsistency using pseudo annotations. In addition, different from the previous works \cite{sun2021dual,chen2021local} which specially design a module to obtain the consistency-related representation, we find that Vision Transformer \cite{dosovitskiy2020image} naturally provides the consistency representation from the attention map among patch embeddings, thanks to their self-attention mechanism. Therefore, we apply it to build the detection network and propose two key components: \textbf{UPCL} (\textbf{U}nsupervised \textbf{P}atch \textbf{C}onsistency \textbf{L}earning) and \textbf{PCWA} (\textbf{P}rogressive \textbf{C}onsistency \textbf{W}eighted \textbf{A}ssemble). 

\textbf{UPCL} is a training strategy for learning the consistency-related representations through an unsupervised forgery location method. We approximately estimate forgery location maps by comparing the Mahalanobis distances between the MVGs of real/fake features and the patch embedding from the middle layer of Vision Transformer (ViT) \cite{dosovitskiy2020image}. During training, forgery location maps are progressively optimized. To model the consistency constraint, we use the Multi-head Attention Map existed in ViT itself as the representation and constrain them in multi-layers for better learning.

\textbf{PCWA} is a feature enhancement module and can take full advantage of the consistency representation through the proposed UPCL module. In details, we utilize the Attention Map between classification embedding and patch embeddings to progressively weighted average the patch embedding of final layer, and concatenate it with classification embedding before feed them into final MLP for forgery classification. The layers providing these Attention Maps are optimized by \textbf{UPCL} for further improvement.

The main contributions of this work are summarized as follows:
\begin{itemize}
\item We propose an unsupervised patch consistency learning strategy based on vision Transformer to make it possible for face forgery detection to focus on intra-frame inconsistency without pixel-level annotations. It greatly improves the generalization of detection without 
additional overhead.
\item We take full advantage of feature representations under the proposed learning strategy to progressively combine global classification features and local patch features, by weighted averaging the latter using the Attention Map between classification embedding and patch embeddings.
\item Extensive experiments demonstrate the superior generalization ability of proposed method and the effectiveness of unsupervised learning strategy.
\end{itemize}

\section{Related Work}
\subsection{Face Forgery Detection}
Early face manipulation methods usually produce obvious artifacts or inconsistencies on generated face images. Such flaws are important cues for early face forgery detection works. For example, Li \textit{et al.}\cite{li2018ictu} observed that the eye blinking frequency of the forgery video is lower than the normal. Later methods extended it to check the inconsistency of 3D head poses to help forgery videos detection\cite{yang2019exposing}.
Similarly, Matern \textit{et al.}\cite{matern2019exploiting} used hand-crafted visual features in eyes, noses, teeth to distinguish the fake faces.

Besides seeking for visual artifacts, frequency clues has also been introduced in forgery detection to improve detection accuracy, such as Two-branch\cite{masi2020two}, F3-Net\cite{qian2020thinking}, FDFL\cite{Li_2021_CVPR}. Meanwhile, attention mechanism proved to be effective in recent studies like Xception+Reg \cite{dang2020detection} and Multi-attention\cite{Zhao_2021_CVPR} \cite{miao2021TBIOM}. Although these methods have achieved perfect performance in the intra-dataset detection, they suffer big performance drop while generalizing to other unseen forgery datasets.

To overcome the difficulties on generalizing to unseen forgeries, works have been done to discover universal properties shared by different forgery methods. Some works focused on inevitable procedures in forgery, such as affine warping (DSP-FWA \cite{li2018exposing}) and blending (Face X-ray \cite{li2020face}). While others observed that certain type of inconsistency exists in different kinds of forgery videos, such as temporal inconsistency (LipForensics \cite{haliassos2021lips}, FTCN+TT \cite{zheng2021exploring}) and intra-frame inconsistency (Local-related \cite{chen2021local}, PCL+I2G \cite{PCL}, Dual \cite{sun2021dual}). However, in order to learn corresponding artifacts or inconsistency cues, Face X-ray\cite{li2020face} and PCL+I2G\cite{PCL} try to generate the large-scale datasets with annotated forged location for their pixel-level supervised learning. The generation process is time consuming and cannot capture the properties of forgery regions. Local-related\cite{chen2021local} and DCL\cite{sun2021dual} try to generate annotated forged location labels by subtracting forgery image with its corresponding real image. However, these paired images are usually unavailable, especially in the real-world scenes. Even though such paired data can be collected, the forgery region annotations tend to be discontinuous and inaccurate, which are sub-optimal for intra-frame consistency supervision. To tackle these issues, we introduce the unsupervised inconsistency-aware method to learn inconsistency cues for general face forgery detection.

\subsection{Transformer}
Transformers \cite{vaswani2017attention} are proposed for machine translation and have become the state of the art method in NLP tasks for their strong ability in modeling long-range context information. Vision Transformer (ViT)\cite{dosovitskiy2020image} adjusted Transformers for computer vision tasks, by modeling image as a sequences of image patches. Several works leveraging transformers to boost face forgery detection have been done: Miao \textit{et al.}\cite{miao2021towards} extend transformer using bag-of-feature to learn local forgery features. Khan \textit{et al.}\cite{khan2021video} propose a video transformer to extract spatial features with the temporal information for detecting forgery. Zheng \textit{et al.}\cite{zheng2021exploring} design a light-weight temporal transformer after their proposed fully temporal convolution network to explore the temporal coherence for general manipulated video detection.
In this paper, we also extend transformer to dig the relationships among different regions and capture more local consistency information for general forgery image detection.

\section{Method}
\begin{figure*}[!t]
    \centering
    \includegraphics[width=0.95\textwidth]{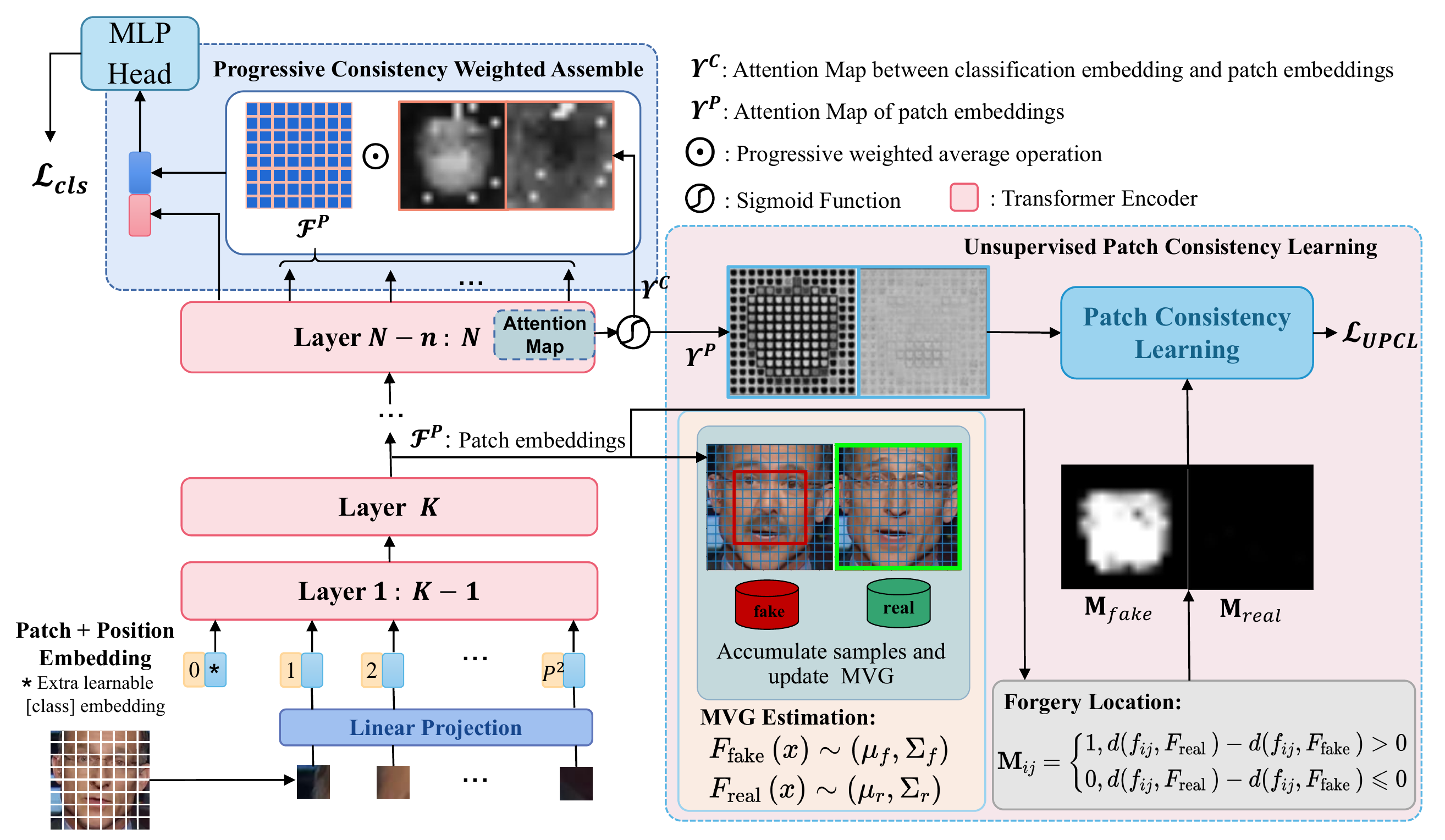}
    \caption{\small{An overview of the proposed UIA-ViT. The UPCL module uses the patch embeddings from layer $K$ to make MVG estimation. The averaged Attention Map from layer $N-n$ to layer $N$ is used for patch consistency learning in UPCL module and as consistency weighted matrix in PCWA module.}}
    \label{framework}
    \vspace{-10pt}
\end{figure*}

In this section, we introduce the details of the proposed Vision Transformer based unsupervised inconsistency-aware face forgery detection. As shown in Fig.\ref{framework}, given an image $I$, our network splits $I$ into fixed-size patches, linearly embeds each of them, adds position embeddings and feeds the resulting sequence into the Transformer encoder. The patch embeddings $\mathcal{F}_P$ from layer $K$ are accumulated for unsupervised approximate forgery location, and the estimated location maps are used for modeling consistency constraint. The Attention Map $\Upsilon_P$ and $\Upsilon_C$ are averaged from layer $N-n$ to layer $N$. $\Upsilon_P$ is used for patch consistency learning, and $\Upsilon_C$ is used as consistency weighted matrix in PCWA. In the end, PCWA outputs consistency-aware feature $F=\textrm{UIA}(I)$, and an MLP head is used to do the final prediction.

\subsection{Unsupervised Patch Consistency Learning}
\subsubsection{Unsupervised Approximate Forgery Location.}

We apply the widely used multivariate Gaussian estimation (MVG) to represent the real/fake image patch features and generate pseudo annotations.
To be concrete, we try to fit a MVG of original image patches and a MVG of forged image patches within \textbf{General Forgery Region (GFR)}. We define GFR as the general manipulated region among different forgery datasets, where the patch features can approximately represent the distribution of actual manipulated face region. Specifically, we designate the GFR as the center square region of the cropped faces.

Assume $x_r$ represents the patch feature from layer $K$ of real sample, and $x_f$ represents the patch feature from layer $K$ within GFR of fake sample.
We model the probability density function (PDF) of $x_r$ using the MVG, defined as:
\begin{equation}
    F_{real}(x_r)=\frac{1}{\sqrt{(2\pi)^D|det\Sigma_r|}}e^{-\frac{1}{2}(x_r-\mu_r)^T\Sigma_r^{-1}(x_r-\mu_r)},
\end{equation}
where $\mu_r \in \mathbb{R}^D$ is the mean vector and $\Sigma_r \in \mathbb{R}^{D\times D}$ is the symmetric covariance matrix of the real distribution. Similarly, the PDF of $x_f$ is defined as $F_{fake}(x_f)$ with mean vector $\mu_f$ and covariance matrix $\Sigma_f$. 

During training, $F_{real}$ and $F_{fake}$ are updated by new ($\mu_r$,$\Sigma_r$) and ($\mu_f$,$\Sigma_f$), which are approximated with the sample mean and sample covariance from the observations ($x_r^1, x_r^2, ...,x_r^n \in \mathbb{R}^{D}$) and ($x_f^1, x_f^2, ...,x_f^n \in \mathbb{R}^{D}$). We accumulate the feature observations from every mini batch of training samples and experimentally update two MVG distributions every 0.5 training epochs.

Given MVG distributions of real and fake, the distances between the patch embeddings $\mathcal{F}_P$ from layer K and MVG distributions are used for forgery location prediction. Assume $f_{ij} \in \mathbb{R}^D$ is the particular feature in position $(i,j)$ of $\mathcal{F}_P$. We adopt Mahalanobis distance for distance measure between $f_{ij}$ and the MVG distributions, defined as
\begin{align}
    d(f_{ij},F_{real})=\sqrt{(f_{ij}-\mu_r)^T \Sigma_r^{-1} (f_{ij}-\mu_r)} ,\\
    d(f_{ij},F_{fake})=\sqrt{(f_{ij}-\mu_f)^T \Sigma_f^{-1} (f_{ij}-\mu_f)} .
\end{align}

Then, for fake samples, the predicted location map $\mathbf{M} \in \mathbb{R}^{P \times P}$ is defined as a binary distance comparison map. The annotation is predicted as 0 when the patch feature is more closed to $F_{real}$ than $F_{fake}$, and otherwise predicted as 1, formalized as:
\begin{equation}
    \mathbf{M}_{ij} = \left\{\begin{matrix}
 1,d(f_{ij},F_{real})-d(f_{ij},F_{fake})>0\\0,d(f_{ij},F_{real})-d(f_{ij},F_{fake})\leqslant 0
\end{matrix}\right..
\label{M}
\end{equation}

Note that location map $\mathbf{M}$ is fixes as the all-zero matrix for real samples.
In particular, in order to guarantee that such patch embeddings from our network capture more local texture information rather than high-level semantic information, 
we perform several visualizations as shown in Fig.\ref{Location Map}, and finally utilize the patch embeddings $\mathcal{F}_P$ from Block 6 of UIA-ViT network (i.e. $K=6$) for approximately estimating forgery location map.

\subsubsection{Patch Consistency Loss.}

In each transformer block of the standard Transformer Encoder, there is a Multi-head Attention layer that firstly calculate the compatibility between queries and keys from different input embeddings, called Attention Map. 
Unlike PCL+I2G\cite{PCL} that specially computes the pair-wise similarity of their extracted feature, our ViT based method directly uses Attention Map from middle layers for self-consistency learning. 

Define the mean Attention Map between different patch embeddings from $N-n$ to $N$ Transformer layers as $\Upsilon^{P} \in \mathbb{R}^{P^2 \times P^2}$, where $P^2$ is the number of patch embeddings. $\Upsilon^P_{(i,j),(k,l)}$ represents the consistency between the embedding in position $(i,j)$ and other patch embedding in position $(k,l)$, and higher value means two positions have higher consistency. With the approximate forgery location map $\mathbf{M}$, we design the consistency loss to supervise the Attention Map, formalized as:
\begin{align}
    & \mathbf{C}_{(i,j),(k,l)} = \left\{\begin{matrix}
    c_1 ,& if & \mathbf{M}_{ij}=0 &and &\mathbf{M}_{kl}=0\\
    c_2 ,& if & \mathbf{M}_{ij}=1 &and &\mathbf{M}_{kl}=1\\
    c_3 ,&   & else
\end{matrix}\right. ,\\
    & \mathcal{L}_{UPCL} = \frac{1}{P^2} \sum_{i,j,k,l}|\textrm{sigmoid}(\Upsilon^P_{(i,j),(k,l)})-\mathbf{C}_{(i,j),(k,l)}|,
\end{align}
where $c_1$,$c_2$, $c_3$ are learnable parameters to avoid instability optimization when MVG estimation is biased in the early training. During training process, we initialize them as (0.6,0.6,0.2), and also optimize $c_1$,$c_2$ to increase and optimize $c_3$ to decrease gradually. After convergence, $c_1$,$c_2$, $c_3$ eventually tend to (0.8, 0.8, 0.0) in our experiments.

\subsection{Progressive Consistency Weighted Assemble}

In order to perform classification, we use the standard approach of adding an extra learnable classification token to the sequence. However, due to final classification embedding capturing more global semantic information rather than local texture, it is not sufficient to utilize the intra-frame inconsistency clues if merely feed the final classification embedding into MLP head. 
Therefore, we propose a novel module, named Progressive Consistency Weighted Assemble (PCWA), which progressively combines global features and local inconsistency-aware patch features for final binary classification. 

Specifically, the patch embeddings of final layer are weighted average with  $\Upsilon^{C} \in \mathbb{R}^{P\times P}$, which is defined as the mean Attention Map between classification embedding and other patch embeddings from $N-n$ to $N$ Transformer layers. Denote classification embedding and patch embeddings of final layer as $\mathcal{F}^{C} \in \mathbb{R}^{D}$, $\mathcal{F}^P \in \mathbb{R}^{P^2 \times D}$.  To avert instability optimization in the early training stage, we adopt variable weight $w$ along with the current iterations. The scalar $w$ is gradually decreased to zero controlled by the decreasing function with hyper-parameters $\rho$ and $\theta$.
Then the weighted matrix $\mathbf{A}^P$ gradually transfers from averaged weighting (all-one matrix) to consistency weighting, formalized as:
\begin{align}
    & w = \textrm{sigmoid}(-\rho (step-\theta)), step = \frac{current\_iters}{total\_iters} \in [0,1], \label{w} \\
    & \mathbf{A}^{P} =w \ast \mathbbm{1} + (1-w)\ast \textrm{sigmoid}(\Upsilon^{C}).
\end{align}

Ideally, the well optimized $\Upsilon^{C}$ can capture consistency information between the input global classification embedding and other local patch embeddings, and suggest which regions should be great considered by the classifier at the end of network. We adopt $\mathbf{A}^P$ as the weighted matrix and apply weighted average operation to patch embeddings $\mathcal{F}^P$. In the end, the average weighted feature $\mathcal{F}^{AP}$ is concatenated with $\mathcal{F}^C$ for final binary classification.
The above procedure can be formulated as:
\begin{align}
    & \mathcal{F}^{AP} = \mathbf{A}^{P} \odot \mathcal{F}^P = \frac{1}{P^2}\sum_{i,j}^{P} \mathbf{A}^P_{ij}*\mathcal{F}^P_{ij}, \\
    & \mathbf{\hat{y}} = \textrm{MLP}(concat\{\mathcal{F}^C, \mathcal{F}^{AP}\}),
\end{align}
where $\mathbf{\hat{y}}$ is the final predicted probability from MLP concatenated after the PCWA module.

\subsection{Loss Functions}
Assume $\mathbf{y}$ represents the binary labels indicating real or fake of input image. We use cross-entropy loss to supervise the final predicted probability $\mathbf{\hat{y}}$ to given binary labels 0/1, defined as:
\begin{equation}
    \mathcal{L}_{cls} = -[\mathbf{y}\log \mathbf{\hat{y}} + (1-\mathbf{y})\log (1-\mathbf{\hat{y}})]. 
\end{equation}
The total loss functions of the proposed method are described as:
\begin{equation}
    \mathcal{L}_{total} = \mathcal{L}_{cls}+\lambda_1 \mathcal{L}_{UPCL}+\lambda_2(\frac{1}{|c1|}+\frac{1}{|c2|})+\lambda_3|c3|,
\label{total loss}
\end{equation}
where the final two losses are used for optimizing the learnable parameters $c_1$, $c_2$ to be increased and $c_3$ to be decreased along with the training process. $\lambda_1$, $\lambda_2$ and $\lambda_3$ are hyper-parameters used to balance the cross-entropy loss and consistency loss, meanwhile adjust the variations of consistency factors $c_1$, $c_2$ and $c_3$.

\section{Experiment}

\subsection{Experimental Setting}
\subsubsection{Datasets.} We conduct the experiments on four forgery datasets: 1) \textbf{FaceForensics++ \cite{rossler2019faceforensics++}} consists of 1,000 original videos and corresponding fake videos which are generated by four manipulation methods: DeepFakes \cite{DeepFakes2019}, Face2Face \cite{thies2016face2face}, FaceSwap \cite{FaceSwap2019}, NeuralTextures \cite{thies2019deferred}. It provides forgery videos of three quality levels (raw, high, low quality). 2)  \textbf{Celeb-DF} \cite{li2020celeb} is tempered by the DeepFake method and has a diversified distribution of scenarios. The author publishes two versions of their dataset, called Celeb-DF(v1) and Celeb-DF(v2). 3) \textbf{DeepFakeDetection(DFD)} \cite{DFD2019} is produced by Google/Jigsaw, which contains 3,068 facial fake videos clips generated from 363 original videos. Forgery videos are tempered by the improved DeepFake method. 4) \textbf{Deepfake Detection Challenge \cite{Seferbekov2020}} (DFDC) preview dataset is generated by two kinds of synthesis methods on 1,131 original videos. Use the stand testset consisted 780 videos for our experiments.

\vspace{-5pt}
\subsubsection{Implement Details.}
Following \cite{thies2016face2face}, we employ the open source dlib algorithm to do face detection and landmark localization. All the detected faces are cropped around the center of the face, and then resized to $224 \times 224$. 
We adopt the ViT-Base architecture \cite{dosovitskiy2020image} as backbone where the input patch size is $16 \times 16$ and the number of encoder layer is set to 12. The designed GFR is assigned as $8 \times 8$ patches in the center region of cropped face.
In the training process, the model is optimized only by cross-entropy loss in the first one epoch, and optimized by the total loss in the next epoch. 
Batch size is set to 96 and the Adam optimizer with the initial learning rate 3e-5 is adopted. The learning rate is reduced when the validation accuracy arrives at plateau. 
For the PCWA module, the hyper-parameters $\rho, \theta$ of variable weight $w$ is set to 12, 0.7, respectively. In loss function Eqn.(\ref{total loss}), we experimentally set the weight $\lambda_1$, $\lambda_2$ and $\lambda_3$ to 0.06, 0.05, 0.5. These hyper-parameters are experimentally explored in supplementary materials.
While testing, we set the $step$ in Eqn.(\ref{w}) to 1.0 and use 110 frames per testing videos following FaceForensics++ \cite{rossler2019faceforensics++}. 

\vspace{-5pt}
\subsection{Quantitative Results}
\subsubsection{Cross Dataset Experiment.}
We first evaluate the detection performance on unseen datasets to demonstrate the generalization ability of our method in Table.\ref{Cross-dataset experimental results}. We train the face forgery detectors on all the four types of fake data in FF++(HQ) and evaluate them on four unseen datasets, including DeepFakeDetection(DFD), Celeb-DF-v2, Celeb-DF-v1 and DFDC preview(DFDC-P).
We report the cross-dataset AUC(\%) in frame level of several state-of-art methods, each of which detects forgery using single frame rather than a video clip.

Table.\ref{Cross-dataset experimental results} shows that the proposed UIA-ViT outperforms other detection methods on several unseen datasets compared with recently general face forgery detection methods. Although our method adopts unsupervised forgery location method, it also superior to \textbf{PCL+I2G}\cite{PCL} and \textbf{Local-relation}\cite{chen2021local} on Celeb-DF, which both devote to extract intra-frame inconsistency cues with forgery location annotation. And we also slightly outperform the newly proposed \textbf{DCL}\cite{sun2021dual}, which designs contrastive learning at different granularities to learn generalized feature representation for face forgery detection.
Except DFDC-P, UIA-ViT achieves the best performance on unseen datasets. For example, UIA-ViT greatly outperforms other methods by 3.0+\% on DFD, demonstrating the effectiveness of our method to improve the generalization ability for face forgery detection.

\begin{table}[t]
\caption{\textbf{Cross-dataset experimental results.} Our models are trained on FaceForensics++ and tested on unseen datasets. We report the video-level AUC(\%) on FaceForensics++(high quality) and frame level AUC(\%) on other testing datasets
}
\small
\centering
\setlength{\tabcolsep}{0.9mm}{
\begin{tabular}{l|c|cccc}
\Xhline{1.0pt}
Methods & FF++.HQ & DFD & Celeb-DF-v2 & Celeb-DF-v1 & DFDC-P\\ \hline \hline
Xception\cite{rossler2019faceforensics++} & 96.30 & 70.47  & 65.50 & 62.33 & 72.20 \\
Capsule\cite{nguyen2019capsule} & 96.46 & 62.75  & 57.50 & 60.49 & 65.95\\
Multi-Attention\cite{Zhao_2021_CVPR} &99.29  & 75.53  & 67.44 & 54.01 & 66.28 \\
FRLM\cite{miao2021TBIOM} & 99.50 & 68.17 & 70.58 & 76.52 & 69.81 \\
Face X-ray\cite{li2020face} & 87.40 & 85.60 & 74.20 & 80.58 & 70.00 \\
LTW\cite{sun2021domain} & 99.17 & 88.56 & 77.14 & —— & 74.58 \\
PCL+I2G\cite{PCL}  & 99.11 & —— & 81.80 &—— &——  \\
Local-relation\cite{chen2021local} & 99.46 & 89.24  & 78.26 & —— & 76.53 \\
DCL\cite{sun2021dual} & 99.30 & 91.66 & 82.30 & ——  & \textbf{76.71} \\ \hline
\textbf{UIA-ViT} & 99.33  & \textbf{94.68} & \textbf{82.41} & \textbf{86.59} & 75.80\\ \Xhline{1.0pt}
\end{tabular}
}
\label{Cross-dataset experimental results}
\end{table}

\begin{table}[t]
\caption{\textbf{Cross manipulation experimental results.} Test on each forgery method dataset (Deepfakes, Face2Face, FaceSwap, NeuralTexture) while training the model on the remaining three datasets} 
\centering
\begin{tabular}{l|cccccccc}
\Xhline{1.0pt}
\multirow{3}{*}{Methods} & \multicolumn{8}{c}{Training on the remaining three forgery dataset} \\ \cline{2-9} 
 & \multicolumn{2}{c|}{Deepfakes} & \multicolumn{2}{c|}{Face2Face} & \multicolumn{2}{c|}{FaceSwap} & \multicolumn{2}{c}{NeuralTextures} \\ \cline{2-9} 
 & \multicolumn{1}{l}{ACC(\%)} & \multicolumn{1}{l|}{AUC(\%)} & \multicolumn{1}{l}{ACC(\%)} & \multicolumn{1}{l|}{AUC(\%)} & \multicolumn{1}{l}{ACC(\%)} & \multicolumn{1}{l|}{AUC(\%)} & \multicolumn{1}{l}{ACC(\%)} & \multicolumn{1}{l}{AUC(\%)} \\ \hline \hline
Xception\cite{rossler2019faceforensics++} & 85.5 & \multicolumn{1}{c|}{92.5} & 77.5 & \multicolumn{1}{c|}{84.5} & 49.3 & \multicolumn{1}{c|}{51.6} & \textbf{70.9} & 77.3 \\
MLDG\cite{li2018learning} & 84.2 & \multicolumn{1}{c|}{91.8} & 63.4 & \multicolumn{1}{c|}{77.1} & 52.7 & \multicolumn{1}{c|}{60.9} & 62.1 & 78.0 \\
LTW\cite{sun2021domain} & 85.6 & \multicolumn{1}{c|}{92.7} & 65.6 & \multicolumn{1}{c|}{80.2} & \textbf{54.9} & \multicolumn{1}{c|}{64.0} & 65.3 & 77.3 \\
DCL\cite{sun2021dual} & 87.7 & \multicolumn{1}{c|}{94.9} & 68.4 & \multicolumn{1}{c|}{82.9} & —— & \multicolumn{1}{c|}{——} & —— & —— \\ \hline
UIA-ViT & \textbf{90.4} & \multicolumn{1}{c|}{\textbf{96.7}} & \textbf{86.4} & \multicolumn{1}{c|}{\textbf{94.2}} & 51.4 & \multicolumn{1}{c|}{\textbf{70.7}} & 60.0  &\textbf{82.8}  \\ \Xhline{1.0pt}
\end{tabular}
\label{cross manipulation}
\vspace{-5pt}
\end{table}

\subsubsection{Cross Manipulation Experiment.}

To assess the generalization ability to unseen manipulations without perturbations such as variations in lighting and facial identities, we conduct the experiments on FaceForensics++ which is consisted of four types of manipulations and the same source videos. We utilize the fake videos created by four different forgery methods: DeepFakes, Face2Face, FaceSwap and NeuralTextures. We evaluate face forgery detectors with the leave-one-out strategy. Specifically, we test on each forgery method data using the model trained on the remaining three forgery methods in the high quality setting (FF++.HQ). 

We compare the proposed method with other state of the art methods, and report the video-level ACC(\%) and AUC(\%). In the Table.\ref{cross manipulation}, there are four compared methods: \textbf{1) Xception}\cite{rossler2019faceforensics++} is trained with official code by ourselves,  \textbf{2) MLDG}\cite{li2018learning} uses meta-learning for domain generalization, which is adapted to generalized face forgery detection in the work \textbf{LTW}\cite{sun2021domain}.
\textbf{3) LTW}\cite{sun2021domain} also uses meta-learning strategy to learn domain-invariant model for unseen domain detection. This work reports the cross-manipulation results of LTW and MLDG. \textbf{4) DCL}\cite{sun2021dual} designs contrastive learning at different granularity to learn generalized feature representation. It reports their cross-manipulation results only on Deepfakes and Face2Face.
The results in Table.\ref{cross manipulation} show that the proposed method consistently achieves superior generalization performance compared to other frame-level methods, especially on Face2Face and FaceSwap with AUC(\%) evaluation metrics. For example, out method achieves 9.7+\% and 6.7+\% higher AUC than other methods, demonstrating the state-of-the-art generalization ability of UIA-ViT to unseen forgeries.

\subsection{Visualization}

\begin{figure*}[!t]
    \centering
    \includegraphics[width=\textwidth]{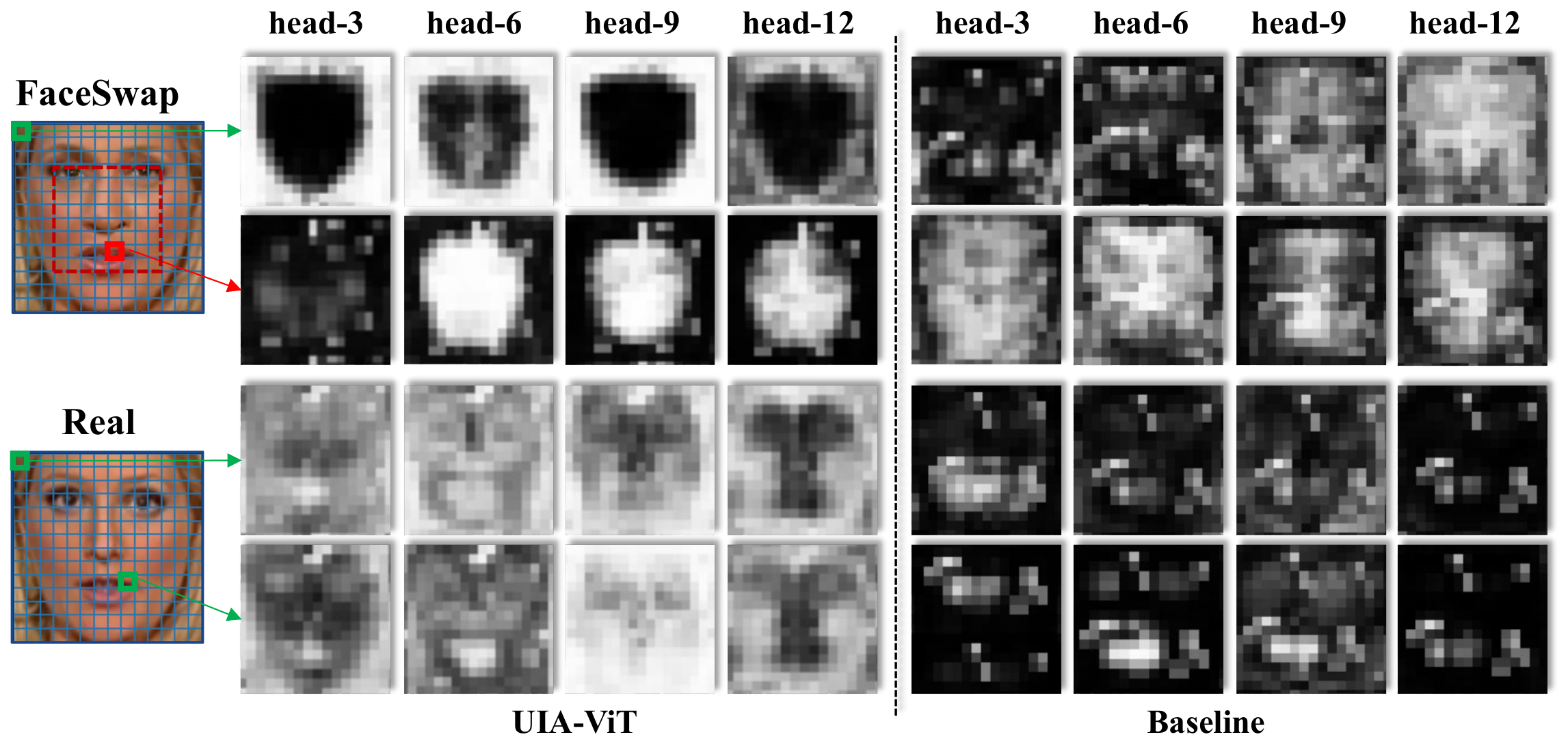}
    \caption{Attention Maps of different queries in head-3/6/9/12 of the 11-th layer. Small squares denotes the query location.}
    \label{visualization}
    \vspace{-8pt}
\end{figure*}

To visually illustrate the consistency-aware embeddings learned by the proposed method, some attention maps of different queries from real or fake samples are shown in Fig.\ref{visualization}. We choose different query locations and show their attention map with keys of all patch embeddings. The dashed red rectangle in forgery sample represents the \textbf{GFR} location where the patch embeddings are used for MVG estimation. 
The query locations are indicated by green and red squares.
From the visualizations, we conclude following observations: 1) patch embeddings among original background of the forged face are similar to each other, and so as those among forgery regions of the forged face. 2) patch embeddings between original background and forgery region is less similar. 3) the similarity between patch embeddings among different locations of real face are relatively equal. Such observations further illustrate the effectiveness of the proposed method to learn the consistency-related representations.

\begin{figure*}[!t]
    \centering
    \includegraphics[width=0.95\textwidth]{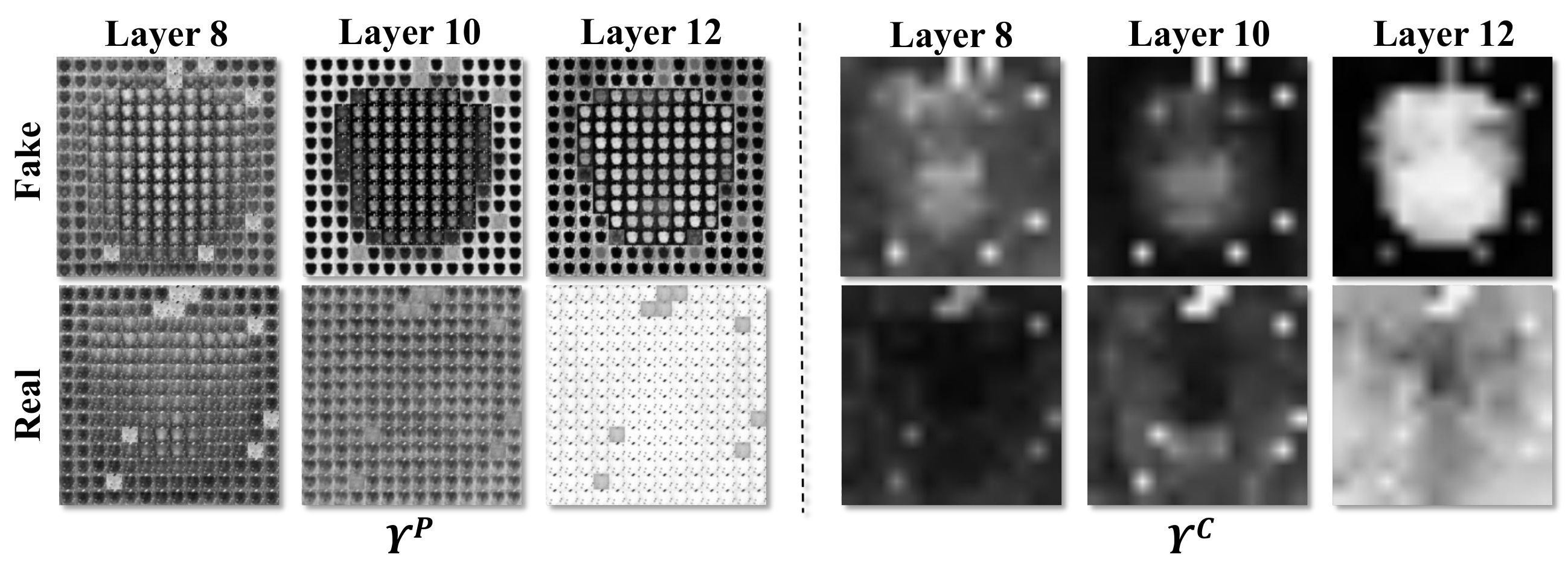}
    \caption{Attention Map of layer 8/10/12 (averaged from multiple heads). The left $\Upsilon^P$ denotes the attention map between patch embeddings. The right $\Upsilon^C$ denotes the attention map between classification embedding and patch embeddings.}
    \label{Attention Map}
\end{figure*}

\begin{figure*}[!t]
    \centering
    \includegraphics[width=\textwidth]{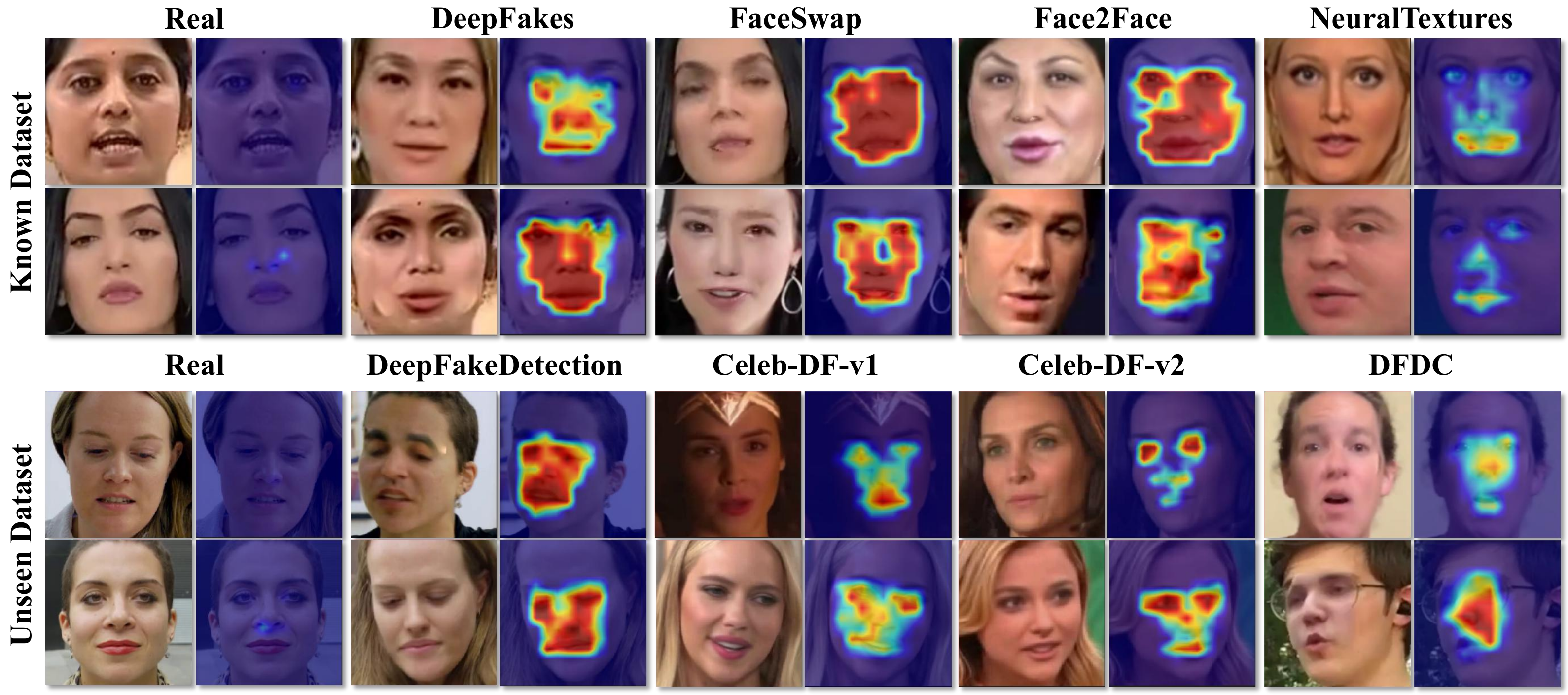}
    \caption{Predicted Location Map of different datasets.}
    \label{Cross Attention Map}
    \vspace{-14pt}
\end{figure*}

Moreover, we show the complete Attention Map (averaged from multiple heads) from middle layers of our model in Fig.\ref{Attention Map}. $\Upsilon^P \in \mathbb{R}^{P^2 \times P^2}$ is the combination of $P^2$ small attention maps, each of which is of the size $P\times P$. $\Upsilon^C \in \mathbb{R}^{P\times P}$ is conducted by calculating the attention map between the classification embedding and the other patch embeddings. The visualizations show the similar observations described in the preceding paragraph. Meanwhile, it shows that Attention Maps in later layer are more closed to our designed consistency constraint.
Though $\Upsilon^C$ has not been constrained by particular attention information, it also contains consistency cues between the input global classification embedding and other local patch embeddings, which indicates the importance regions should be great considered by the final classifier.

More \textbf{Predicted Location Map} of different datasets are further shown in Fig.\ref{Cross Attention Map}. The UIA-ViT model is trained on FaceForensics++ consisted of four manipulation datasets, Deepfakes, FaceSwap, Face2Face, NeuralTextures and original videos (Real). We observe that our model can well concentrate on the manipulated regions, e.g. DF, F2F, FS replace the most areas of source faces with manipulated target faces, and NT manipulates the low-half faces, mainly on mouth and nose regions. 
We also show the predicted location maps of other unseen datasets, including DeepFakeDetection, Celeb-DF-v1, Celeb-DF-v2 and DFDC. They all use face swapping methods, which manipulate the identity of source faces and retain the original background. Their forgery regions should locate in center area of faces, like eyes, nose and mouth. When generalizing to these unseen datasets, we can find that most predicted forgery regions are consistent with the speculation.

\begin{table}[!t]
\caption{Ablation Study for the effect of different components. All models are trained in FaceForensics++ and tested on the unseen forgery dataset Celeb-DF. The default components are expressed in bold}
\centering

\setlength{\tabcolsep}{0.8mm}{
\begin{tabular}{ccc|cc|c|c}
\Xhline{1.0pt}
PCL & UPCL-hard & \textbf{UPCL} & CWA & \textbf{PCWA} & Celeb-DF-v1 & Celeb-DF-v2 \\ \hline \hline
- & - & - & - & - & 75.32 & 76.25 \\ \hline
\checkmark &  &  &  &  & 77.88 & 79.85 \\ \hline
 & \checkmark &  &  &  & 78.54 & 78.55 \\ \hline
 &  & \checkmark &  &  & 82.96  & 80.86 \\ \hline
 &  & \checkmark & \checkmark &   & 84.77 & 81.77 \\ \hline
 &  & \checkmark &  & \checkmark & \textbf{86.59} & \textbf{82.41} \\ \Xhline{1.0pt}
\end{tabular}}
\label{Ablation Study}

\end{table}

\subsection{Ablation Study}
To explore the effectiveness of different components of the proposed UIA-ViT, we spilt each part separately for verification. Specifically, we develop the following experiment comparisons: 
\textbf{1) baseline}: ViT-Base with the same training details. \textbf{2) PCL}:  Patch consistency learning which supervised by fixed GRF as general location map. \textbf{3) UPCL-hard}: Unsupervised patch consistency learning module, but fix consistency factors $c_1$=1,$c_2$=1 and $c_3$=0.  \textbf{4) UPCL}: Stand unsupervised patch consistency learning module. \textbf{5) CWA}: Consistency Weighted Assemble which fixes progressive weighted function as zero in Eqn.(\ref{w}). \textbf{6) PCWA}: Stand progressive consistency weighted assemble module.

The experimental results are shown in Table.\ref{Ablation Study}, and all models are trained on FaceForensics++ and tested on the unseen forgery dataset Celeb-DF.
We draw the following conclusions: 1) Compared with \textbf{UPCL}, the model equipped with \textbf{PCL} and \textbf{UPCL-hard} are less generalized to Celeb-DF.
Because MVG estimation in UPCL can help to amend the prior GFR and generate the predicted location map more closed to correct manipulated region. And soft consistency factors guarantee stable learning in early stage of training. 
2) Both \textbf{CWA} and \textbf{PCWA} further improve the performance based on \textbf{UPCL}, demonstrating the importance of Attention Map $\Upsilon^C$ which indicates noteworthy regions for final classifier. Between them, \textbf{PCWA} performs better because of the progressive mechanism.
3) Comparing the results of \textbf{baseline}, \textbf{UPCL} and \textbf{UPCL+PCWA}, it demonstrates that both of proposed modules can improve the forgery detection performance. The model equipped with both components can obtain further improvements by 11\% on Celeb-DF-v1 comparing to baseline, which demonstrates that the two components are complementary to each other.



\begin{figure*}[!t]

\begin{minipage}[!t]{\textwidth}
	\includegraphics[width=\textwidth]{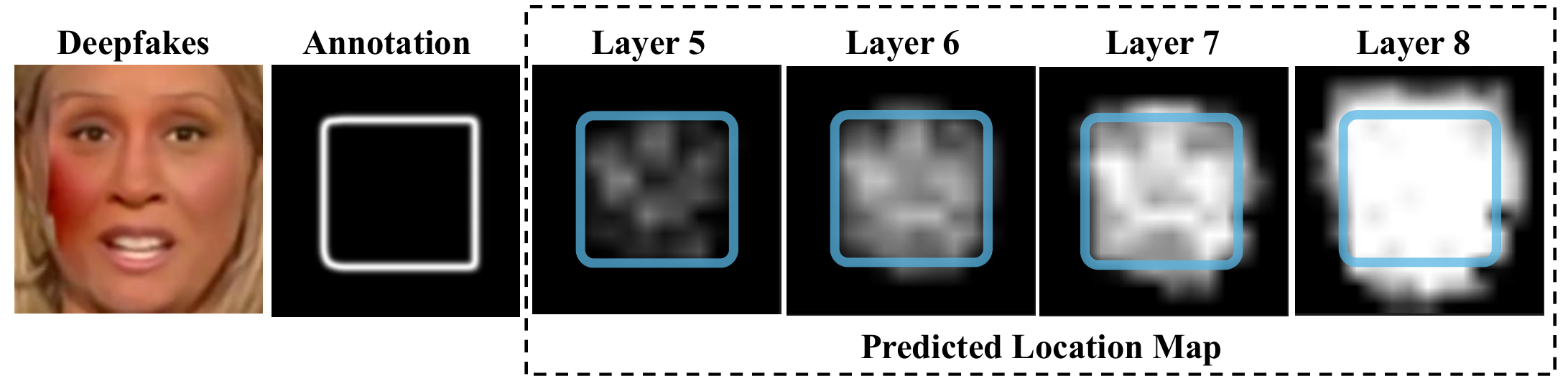}
	\caption{Predicted Location Map of different layers.}
	\label{Location Map}
\end{minipage}

\begin{minipage}[!t]{0.6\textwidth}
	\makeatletter\def\@captype{table}\makeatother
	\caption{Cross-dataset AUC of different GFR}
	\vspace{3pt}
	\setlength{\tabcolsep}{0.5mm}{
	\begin{tabular}{l|cccc}
	\Xhline{1.0pt}
	AUC(\%) & GFR low & GFR large & GFR small & GFR \\ \hline
	Celeb-DF-v2 & 79.68 & 79.33 & 79.04  & \textbf{82.41} \\ 
	DFDC-P & 74.61 & 73.78 & 73.13 & \textbf{75.80} \\ \Xhline{1.0pt}
	\end{tabular}
	}

	\subfigure{\includegraphics[width=\textwidth]{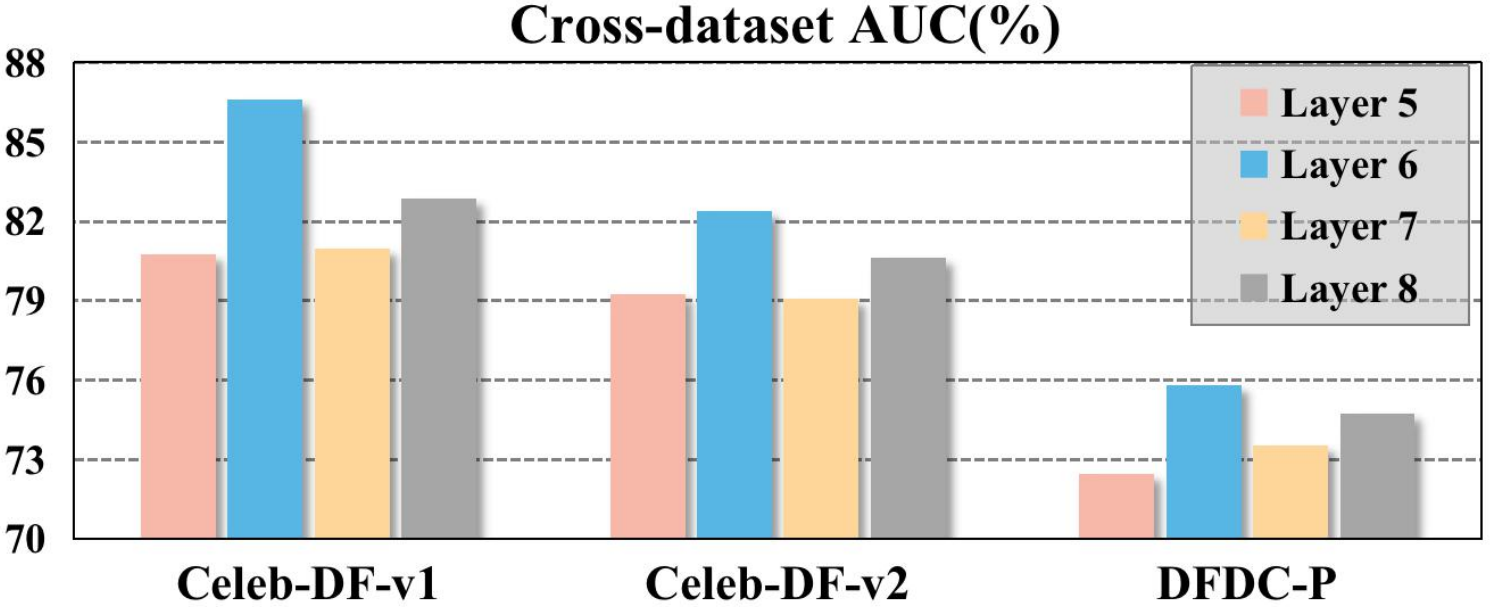}}
	\caption{Compare cross-dataset AUC of Layer 5-8.}
	\label{Location Map quantitation}

\label{GFR experiment}
\end{minipage}
\hspace{3pt}
\begin{minipage}[t]{0.37\textwidth}
	\vspace{-90pt}
	\subfigure{\includegraphics[width=\textwidth]{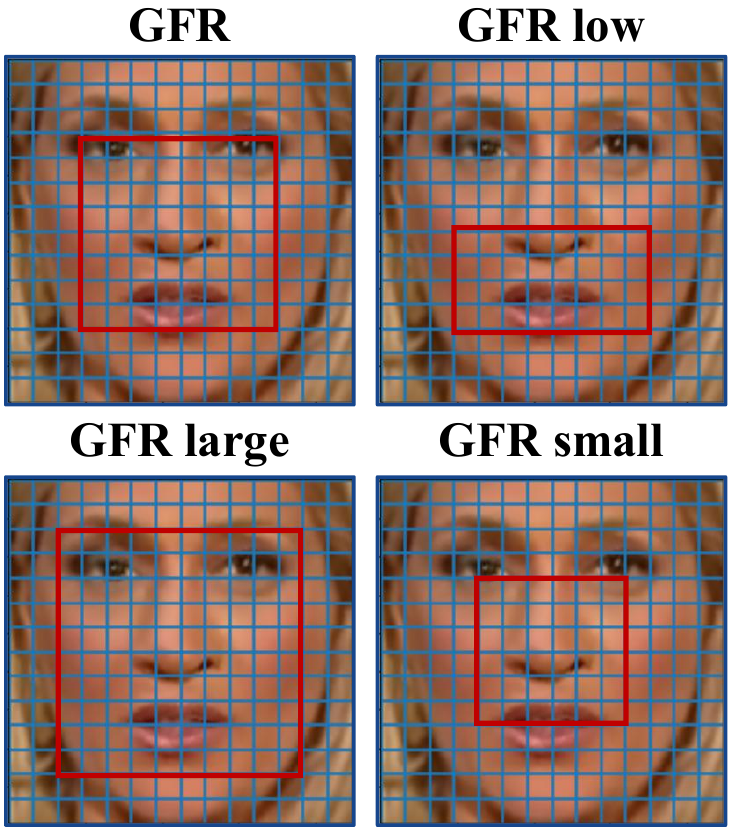}}
	\caption{Different GFR.}
	\label{GFR}
\end{minipage}

\end{figure*}

\subsection{Analysis}
\subsubsection{Determine Which Layer for MVG Estimation.}
In order to determine which layer is the best choice to conduct MVG estimation for forgery location prediction, we train the baseline ViT-Base model, and extract the patch embeddings from different middle layers for updating corresponding MVG distributions. Then, we  visualize the predicted forgery location maps estimated by MVG distributions from 5-th, 6-th, 7-th and 8-th layers. Notice that different from the binary operation in Eqn.(\ref{M}), here predicted location map is computed as $\mathbf{M}_{ij}=\textrm{ReLU}(d(f_{ij},F_{real})-d(f_{ij},F_{fake}))$, where non-zero value indicates predicting as fake.

The visualizations are shown in Fig.\ref{Location Map}. The annotation of forgery location is produced using the mask generation method mentioned in Face X-ray\cite{li2020face} which delimits the foreground manipulated region of forged image. We observe that: 1) the distances between real and fake MVG distributions are larger in the later layers; 2) the extracted features in the foreground and background are also more distinguishable in the later layers; 3) the predicted location map gradually expands to the whole image, because features in later layers capture more high-level semantic information rather than local texture information. 
Among them, we find that the predicted location map from 6-th layer is more closed to the annotation and apply it to conduct MVG estimation for forgery location prediction.
We further make the quantitative analysis on this issue, as shown in Fig.\ref{Location Map quantitation}. The experimental results show that utilizing the embeddings of 6-th layer to conduct unsupervised forgery location can achieve better generalization performance on several unseen datasets.

\subsubsection{Determine Which GFR for MVG Estimation.}
We further explore the effect of different locations and sizes of General Forgery Region (GFR), and develop following experiments: \textbf{GFR low}, \textbf{GFR large}, \textbf{GFR small}, as shown in Fig.\ref{GFR}. GFR low is assigned as the low-half face, where nearly all pixels are manipulated in FaceForensics++ but eye region is missing. GFR large is assigned as the bigger region which contains most manipulated region and several mistake real pixels. Relatively, GFR small contains less manipulated region and less mistake real pixels. GFR is a trade-off proposal that locates in the center square region of faces, which covers most manipulated region of FaceForensics++. 
Comparing the performance of different locations and sizes of GFR in Table.\ref{GFR experiment}, we find that the standard GFR is superior than others. It further demonstrates that the features within GFR in FaceForensics++ can represent the distribution of actual manipulated face region and thus generalize well to other unseen datasets.

\section{Conclusions}
In this paper, we propose a novel face forgery detector named UIA-ViT, which is based on Vision Transformer and equipped with two key components UPCL and PCWA. UPCL is a training stragety for learning the consistency-related representations through an unsupervised forgery location method.
PCWA is a feature enhancement module and can take full advantage of the consistency representation. Visualizations show the great capabilities on learning consistency-related representations of our proposed method. Extensive experiments evidence effectiveness of our method for general face forgery detection.


\subsubsection{Acknowledgements} This work is supported by the National Natural Science Foundation of China (No. 62002336, No. U20B2047) and Exploration Fund Project of University of Science and Technology of China under Grant YD3480002001.

\clearpage
%
%

\end{document}